\documentclass[letterpaper, 10 pt, conference]{ieeeconf}
\IEEEoverridecommandlockouts

\usepackage{cite}

\usepackage{amsthm}
\usepackage{amsmath,amssymb,amsfonts,bm,mathrsfs}
\usepackage{graphicx}
\usepackage[hidelinks]{hyperref}
\usepackage{cleveref}
\usepackage{tikz}
\usepackage{algorithm}
\usepackage{algpseudocode}
\usepackage[dvipsnames]{xcolor}
\usetikzlibrary{arrows.meta,positioning,calc,intersections,fit,backgrounds}
\usepackage[caption=false,font=footnotesize]{subfig}
\usepackage{microtype}
\usepackage{orcidlink}
\usepackage{booktabs}
\usepackage{siunitx}

\allowdisplaybreaks
\newcommand{\ssymbol}[1]{^{\@fnsymbol{#1}}}

\title{\Huge Physics-Constrained Denoising Autoencoders for Data-Scarce Wildfire UAV Sensing}

\author{Abdelrahman~Ramadan\textsuperscript{1},
        Zahra~Dorbeigi~Namaghi\textsuperscript{2},
        Emily~Taylor\textsuperscript{2},
        Lucas~Edwards\textsuperscript{2},\\
        Xan~Giuliani\textsuperscript{2},
        David~S.~McLagan\textsuperscript{3},
        Sidney~Givigi\textsuperscript{4},
        Melissa Greeff\textsuperscript{1}%
        \thanks{\textsuperscript{1}A.~Ramadan, M.~Greeff are Graduate student/ Asst. Professor with Electrical and Computer Engineering, Smith Engineering, and with Ingenuity Labs Research Institute, Queen's University, Kingston, ON K7L 3N6, Canada (e-mails: (20amr3, melissa.greef)@queensu.ca ).}%
        \thanks{\textsuperscript{2}Z.~Dorbeigi~Namaghi,  E.~Taylor, L.~Edwards, and X. Giuliani are the graduate researcher/research assistant/technical team with FEWALab and Geological Sciences and Geological Engineering, Queen's University, Kingston, ON K7L 3N6, Canada (e-mails: 24tcx2@queensu.ca, 22MCW3@queensu.ca, 22MH54@queensu.ca, e.taylor@queensu.ca, respectively).}%
        \thanks{\textsuperscript{3}D. S. McLagan is PI of the FEWALab, Affiliate Member of Ingenuity Labs Research Institute, and Asst. Professor with the School of Environmental Studies; Geological Sciences and Geological Engineering, Queen's University, Kingston, ON K7L 3N6, Canada.}%
        \thanks{\textsuperscript{4}S.~Givigi is a Professor with the School of Computing and with Ingenuity Labs Research Institute, Queen's University, Kingston, ON K7L 3N6, Canada (e-mail: sidney.givigi@queensu.ca).}%
}

\begin{document}
\maketitle

\begin{abstract}
Wildfire monitoring requires high-resolution atmospheric measurements, yet low-cost sensors on Unmanned Aerial Vehicles (UAVs) exhibit baseline drift, cross-sensitivity, and response lag that corrupt concentration estimates. Traditional deep learning denoising approaches demand large datasets impractical to obtain from limited UAV flight campaigns. We present PC$^2$DAE, a physics-informed denoising autoencoder that addresses data scarcity by embedding physical constraints directly into the network architecture. Non-negative concentration estimates are enforced via softplus activations and physically plausible temporal smoothing, ensuring outputs are physically admissible by construction rather than relying on loss function penalties. The architecture employs hierarchical decoder heads for Black Carbon, Gas, and CO$_2$ sensor families, with two variants: PC$^2$DAE-Lean (21k parameters) for edge deployment and PC$^2$DAE-Wide (204k parameters) for offline processing. We evaluate on 7,894 synchronized 1 Hz samples collected from UAV flights during prescribed burns in Saskatchewan, Canada (approximately 2.2 hours of flight data), two orders of magnitude below typical deep learning requirements. PC$^2$DAE-Lean achieves 67.3\% smoothness improvement and 90.7\% high-frequency noise reduction with zero physics violations. Five baselines (LSTM-AE, U-Net, Transformer, CBDAE, DeSpaWN) produce 15--23\% negative outputs. The lean variant outperforms wide (+5.6\% smoothness), suggesting reduced capacity with strong inductive bias prevents overfitting in data-scarce regimes. Training completes in under 65 seconds on consumer hardware.
\end{abstract}


\section{Introduction}

Environmental sensing from mobile platforms presents a fundamental signal processing challenge: raw sensor outputs are corrupted by noise, drift, and cross-sensitivity artifacts that obscure the underlying physical quantities of interest. The goal is to recover clean time series that accurately represent pollutant concentrations from noisy observations, while ensuring the denoised outputs remain physically valid (e.g., non-negative concentrations). This paper addresses the specific challenge of achieving reliable denoising when training data is severely limited, as is typical in field campaigns where only hours of labeled data may be available.

\begin{figure}[!t]
    \centering
    \includegraphics[width=0.9\linewidth]{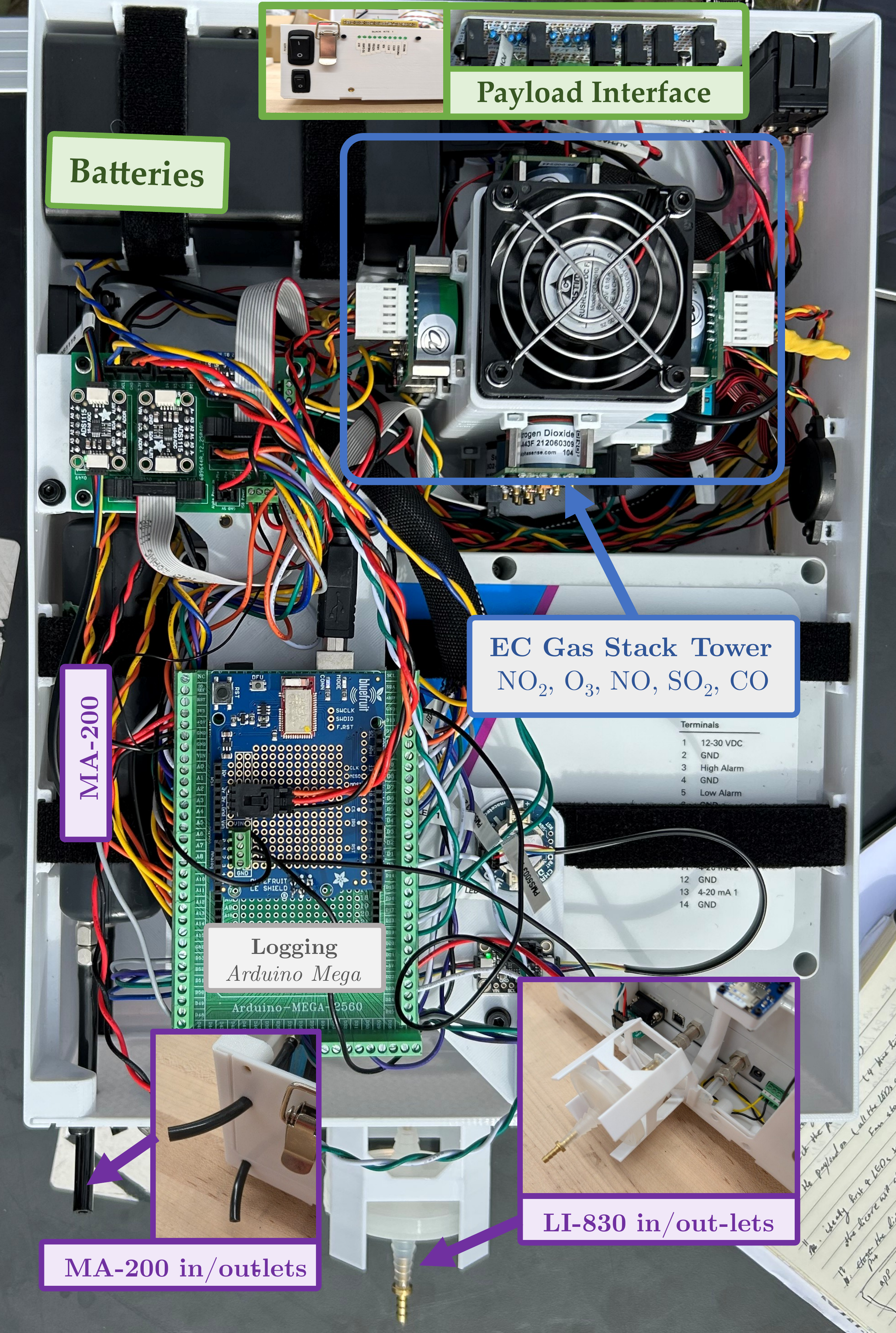}
    \caption{The Black Kite sensor suite mounted on an Aurelia X6 Pro V2 hexacopter. The payload integrates 15 sensors across five families (Black Carbon, Gas, PM, CO$_2$, Environmental) sampling at 1\,Hz, producing synchronized multi-channel time series for post-processing. This platform provides the data used to develop and evaluate our denoising architecture.}
    \label{fig:blackkite}
\end{figure}

The intensification of wildfire regimes across North America, exemplified by the 2023 Canadian wildfire season, which consumed $\approx$ 16 Mha (million hectares, approximately 600\% above the decadal mean), has exposed fundamental limitations in existing environmental monitoring infrastructure~\cite{chen_auto-encoders_2023, heNetworkLowcostAir2022}. Fixed Federal Reference Method (FRM) and Federal Equivalent Method (FEM) stations provide traceable measurements but exhibit sparse spatial coverage (often $>$50\,km spacing in fire-prone regions) and cannot reposition during rapidly evolving events. Satellite retrievals offer synoptic coverage but suffer from coarse spatial resolution (250\,m--1\,km pixels), limited temporal cadence for polar orbiters, and degraded retrieval fidelity under heavy aerosol loading~\cite{usepaWildfireSmokeAir2021, bittner_performance_2024}.

Small Unmanned Aerial Vehicles (UAVs) equipped with compact pollutant sensors present a complementary observing modality capable of targeted in-situ sampling along prescribed trajectories through evolving plumes. Airborne platforms operate under strict Size, Weight, and Power (SWaP) constraints, with payload mass and power budgets typically limited to $<$5\,kg and $<$5\,W for 20--30 minute missions. These limitations rule out reference-grade instruments such as chemiluminescence NO$_x$ analyzers, cavity ring-down spectrometers, and research-grade aethalometers (e.g., Magee Scientific AE33), and instead require lightweight Electro-Chemical (EC) sensors (e.g., Alphasense A4-series) and optical particle counters (e.g., Plantower PMS5003), which are known to exhibit reduced measurement fidelity~\cite{nalakurthi_challenges_2024, papaconstantinouFieldEvaluationLowcost2023a}. Specifically, EC sensors suffer from baseline drift exceeding 40\% over multi-month deployments, cross-sensitivity to interfering gasses (e.g., O$_3$ affecting NO$_2$ response), Arrhenius-mediated temperature dependence, and response lag on the order of tens of seconds, artifacts that necessitate post-processing correction.

This work focuses on the post-processing denoising system rather than the sensing platform itself. The Black Kite system (Figure~\ref{fig:blackkite}) serves as the data source informing our architectural decisions, but the primary contribution is a generalizable physics-constrained denoising architecture applicable to any multi-channel sensor time series where physical constraints are known.

\subsection{The Data Scarcity Challenge}

The fundamental constraint shaping our architectural choices is statistical: the entire labeled dataset comprises approximately $8\times10^{3}$ samples at 1\,Hz cadence, $\sim$2 hours of cumulative flight time. While we expect this number to develop as the project advances, this sample budget is orders of magnitude below the data requirements of conventional deep learning architectures, which typically demand $10^5$--$10^7$ samples for reliable generalization in comparable signal processing tasks.

The 1\,Hz sampling rate introduces additional complexity. While manufacturer specifications indicate some of the EC sensor (Alphasense) response times ($t_{90}$) of 25--80\,s, characterization of data from the integrated Black Kite system reveals substantially faster response times ($t_{90} \approx 4$\,s median, 3--12\,s interquartile range), likely due to the optimized flow-through chamber design. This reduces effective spatial blurring from a theoretical 400\,m (at manufacturer specs) to approximately 7\,m at the mean UAV speed of 1.78\,m/s (GPS and EKF estimates agree within 0.4\%; up to 34\,m at the maximum observed speed of 8.6\,m/s). Flight analysis indicates that 63.1\% of flight time was in active motion (speed $>$0.1\,m/s) versus 36.9\% stationary.

\subsection{Physics as Architectural Constraint}

This data scarcity mandates a design philosophy where physics constraints substitute for model capacity. Rather than deploying high-capacity architectures and relying on regularization to prevent overfitting, we engineer minimal representational capacity constrained by known physics. Each physical constraint, positivity, mass conservation, and spectral consistency, eliminates vast regions of hypothesis space that would otherwise require exponentially more data to rule out empirically.

The resulting architecture, Physics-Constrained Parsimonious Channel-wise Denoising Autoencoder (PC$^2$DAE), achieves reliable denoising in two variants: PC$^2$DAE-Lean with approximately 21,000 trainable parameters optimized for edge deployment, and PC$^2$DAE-Wide with approximately 204,000 parameters for scenarios where computational resources permit larger models. The architecture is currently executed offboard for post-flight processing but is designed with edge deployment in mind, enabling future onboard real-time denoising on UAV-class hardware.

\subsection{Contributions}

The contributions of this work are:
\begin{enumerate}
    \item A hierarchical physics-constrained autoencoder (PC$^2$DAE) with family-specific decoder heads embedding sensor physics for Black Carbon (BC), a collection of NO, NO$_2$, O$_3$, SO$_2$, and CO henceforth referred to as ``Gas'', and CO$_2$ channels, provided in two architecture variants (Lean: 21k params, Wide: 204k params).
    
    \item Learnable temporal smoothing modules with per-channel kernels and adaptive blending that preserve physics constraints while enabling channel-specific noise suppression.
    
    \item Comprehensive evaluation against five deep learning baselines demonstrating state-of-the-art denoising with zero physics violations, where unconstrained baselines produce 15--23\% physically inconsistent outputs.
\end{enumerate}

\section{Related Work}

\subsection{Calibration of Low-Cost Environmental Sensors}

Low-cost EC and optical sensors present fundamental challenges arising from their electrochemical operating principles~\cite{nalakurthi_challenges_2024}. The amperometric mechanism exhibits Arrhenius-mediated temperature sensitivity (approximately doubling the reaction rate per 10$^\circ$C), while the aqueous electrolyte demonstrates hygroscopic behaviour under humidity cycling. Field deployments document baseline drift exceeding 40\% over six-month periods, cross-sensitivity matrix variations up to 300\% beyond specification, and sensor-to-sensor variability reaching 30\% among nominally identical units~\cite{nalakurthi_challenges_2024}. Nonetheless, preliminary data collected by the system are indicative of strong correlations between primary pollutants measured by EC and optical sensors, which demonstrate the potential of the application. Moving forward, it will be critical to develop more effective sensor/instrument calibrations and uncertainty testing.

Machine learning calibration approaches including random forests, gradient boosting, and neural regressors achieve a coefficient of determination $R^2$ of 0.85-0.92 under benign conditions but degrade to 0.4-0.6 in extreme environments~\cite{nalakurthi_challenges_2024}. Critically, these approaches treat the sensor as a black box, learning purely empirical mappings without exploiting known physics. The resulting models require extensive co-location data that is unavailable in wildfire monitoring contexts.

Denoising autoencoders learn to reconstruct clean signals from corrupted observations by forcing robust latent representations~\cite{chen_auto-encoders_2023}. Recent advances include adversarial training, achieving denoising at a Signal-to-Noise Ratio (SNR) as low as $-7$\,dB~\cite{TargetedAdversarialDenoising}, variational formulations with Wasserstein objectives addressing posterior collapse~\cite{tolstikhinWassersteinAutoEncoders2018}, and masked pretraining handling severely sparse inputs~\cite{mcevoyPhysicsInformedMaskedAutoencoder2024}. For sequential data, temporal convolutional architectures offer advantages over recurrent networks: stable gradient propagation, parallelizable training, and flexible receptive field control through dilated convolutions. However, standard architectures require $10^5$--$10^6$ samples for reliable training, two orders of magnitude beyond our data budget. Physics-informed neural networks (PINNs) incorporate domain knowledge through loss function penalties on governing equation residuals~\cite{vogiatzoglouPhysicsinformedNeuralNetworks2025, jeongAdvancedPhysicsinformedNeural2025}. Mechanics-informed autoencoders embed physical relationships into architecture~\cite{liMechanicsinformedAutoencoderEnables2024}. Self-supervised methods combined with physics constraints achieve performance approaching supervised learning with 10$\times$ less labeled data~\cite{hayounPhysicsSemanticInformed2024, lai_physics-informed_2024}.

\section{Sensing Platform and Data Characteristics}

This section describes the data collection platform and signal characteristics that inform our denoising architecture design. The Black Kite system provides the multi-channel sensor time series used for development and evaluation.

\subsection{Platform Architecture}

The Black Kite system (Figure~\ref{fig:blackkite}) employs the Aurelia X6 Pro V2 hexacopter (6\,kg payload capacity). The sensor payload ($\sim$4\,kg) comprises: LI-830 CO$_2$ Non-Dispersive Infrared (NDIR) analyzer (reference-grade), MA200 microAeth 5-wavelength aethalometer (UV 375nm, Blue 470nm, Green 528nm, Red 625nm, IR 880nm), internal sensor tower with Alphasense A4-series EC cells, dual PMS5003 optical particle counters, Sensirion SCD30 NDIR CO$_2$ sensor, and BME280 environmental monitor.

\subsection{The 1\,Hz Sampling Constraint}

All sensor streams are synchronized to 1\,Hz cadence. This rate introduces three critical constraints:

Nyquist aliasing of rapid transients: The 0.5\,Hz Nyquist limit aliases concentration changes faster than 2\,s. Plume boundary crossings at observed UAV speeds (mean 1.78\,m/s, median 0.21\,m/s, max 8.6\,m/s) produce concentration gradients changing over 1--10\,m spatially, corresponding to 0.1--5.6\,s temporally, with aliasing primarily during higher-speed transects.

Heterogeneous sub-Nyquist sampling: The SCD30 CO$_2$ sensor reports at 0.5\,Hz; every alternate logged sample is a repeat of the previous measurement (flagged as ``stale''). The logger propagates per-channel staleness flags, enabling appropriate loss weighting during training.

Sparse temporal supervision: At 1\,Hz, 7,894 samples represent $\sim$2.2 hours of flight data. Raw sensor signals exhibit significant noise, on average, across all channels, 30.8\% of values violate positivity constraints before processing, motivating physics-constrained denoising. Missing data affects $<$4\% of samples.

\begin{figure}[!t]
\centering
\includegraphics[width=0.9\columnwidth]{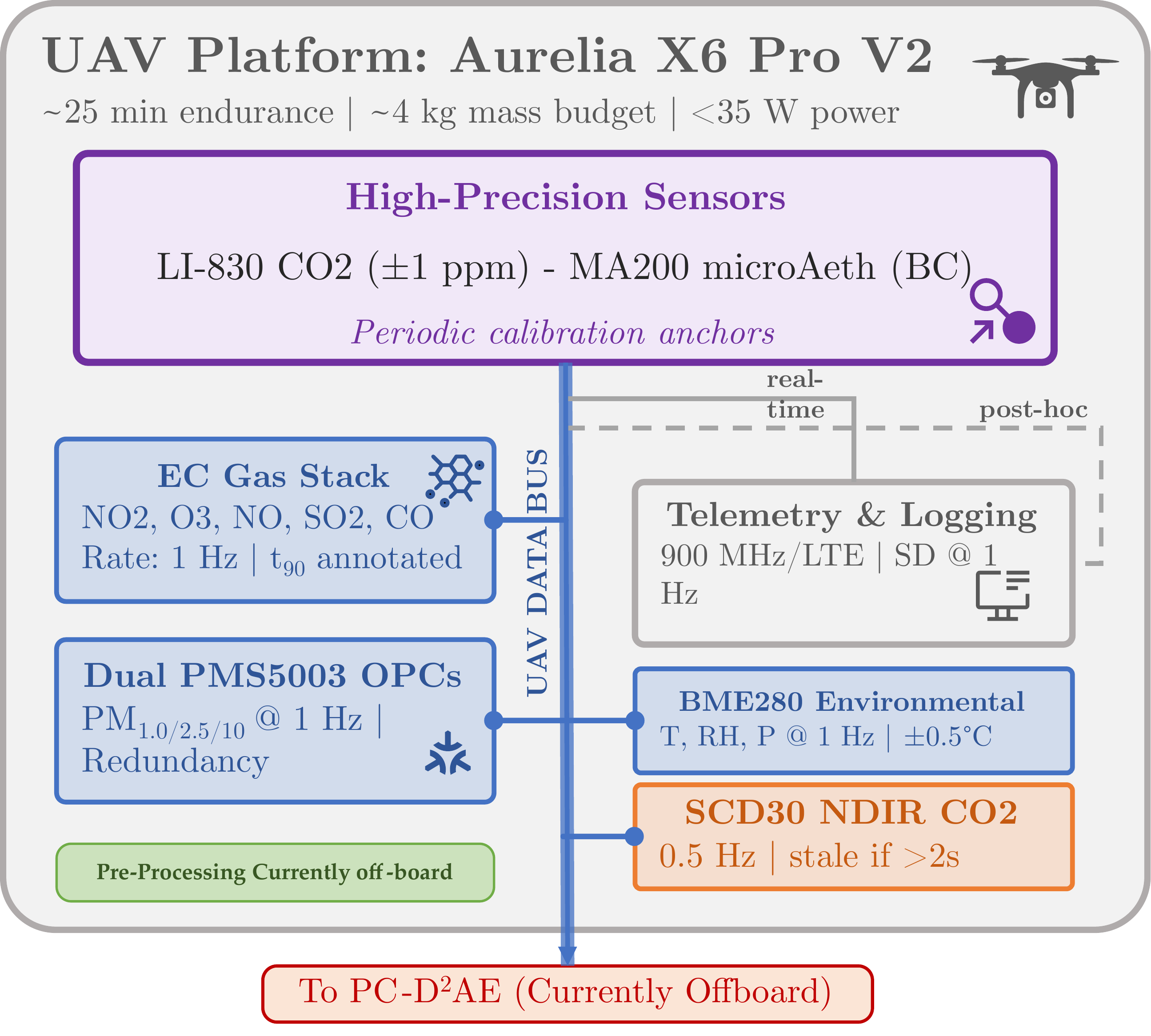}
\caption{Data flow architecture showing multi-sensor integration and preprocessing. All streams are synchronized to 1\,Hz for post-flight denoising via PC$^2$DAE, which currently operates offboard but is designed for future onboard deployment.}
\label{fig:sensor_arch}
\end{figure}

\subsection{Electrochemical Signal Model}

EC channels provide Working Electrode (WE) and Auxiliary Electrode (AE) voltages. Define effective voltages $\mathrm{WE}_e := \mathrm{WE} - \mathrm{WE}_0$, $\mathrm{AE}_e := \mathrm{AE} - \mathrm{AE}_0$ relative to factory zeros ($\mathrm{WE}_0$, $\mathrm{AE}_0$). The AAN-803 compensation template~\cite{ApplicationNotes} yields the compensated signal $s_i$ for channel $i$:
\begin{equation}
\begin{aligned}
    s_i \;&=\; \underbrace{\kappa_{i,0}+\kappa_{i,1}(T-T_0)}_{\text{temperature baseline}} \\ &+ \underbrace{\beta_{i}(T)\bigl(\mathrm{WE}_{e}-n_{i}(T)\,\mathrm{AE}_{e}\bigr)}_{\text{gain \& auxiliary subtraction}},
\label{eq:ec_comp}
\end{aligned}
\end{equation}
where $T$ is temperature, $T_0$ is reference temperature, $\kappa_{i,0}$ and $\kappa_{i,1}$ are baseline offset and temperature coefficient respectively, and $\beta_i(T)$, $n_i(T)$ are temperature-dependent gain and auxiliary subtraction factors. This physics model informs our environmental conditioning module.

\subsection{Dataset and Sensor Family Selection}

Evaluation uses 7,894 synchronized samples from prescribed burn campaign (Saskatchewan, October 2025). Target channels span three sensor families, totaling 23 channels:

Black Carbon (BC, 4 channels): UV and IR wavelengths (BC1, BC2) from the internal aethalometer providing the primary smoke tracer with the highest signal-to-noise ratio.

Gas (9 channels): NO, NO$_2$, O$_3$, SO$_2$, and CO from dual Alphasense Analog Front End (AFE) boards, capturing combustion gas signatures and photochemical processing.

CO$_2$ (2 channels): SCD30 and LI-830 NDIR sensors provide reference-grade validation and low-cost comparison.

Excluded families: Particulate Matter (PM) channels were excluded due to 66.8\% below-detection-limit values from PMS5003 sensors ($\sim$5--10 $\mu$g/m$^3$ detection threshold). Brown Carbon (BrC) was excluded due to insufficient signal-to-noise ratio in the multi-wavelength aethalometer data for reliable BrC quantification.

Environmental conditions spanned 15--35$^\circ$C, 30--70\% RH, with BC concentrations ranging from background ($<$500\,ng/m$^3$) to heavy smoke ($>$50,000\,ng/m$^3$).

\section{Physics-Constrained Hierarchical Architecture}

We present PC$^2$DAE, a hierarchical architecture that embeds domain-specific physical constraints directly into the network structure. The architecture is provided in two variants: PC$^2$DAE-Lean (21k parameters) optimized for edge deployment on resource-constrained UAV platforms, and PC$^2$DAE-Wide (204k parameters) for scenarios permitting larger computational budgets. Both variants employ family-specific decoder heads for BC, Gas, and CO$_2$, where physics constraints serve as the primary mechanism for valid output generation.

\subsection{Constraint-Induced Sample Efficiency}

The denoising task maps $\mathbb{R}^{T \times C} \to \mathbb{R}^{T \times C}$, where $T$ denotes the temporal window size and $C$ the number of channels. With window size $T=128$ samples and $C=23$ channels across three sensor families, this yields $>$2,900 output dimensions per sample. Standard statistical learning bounds suggest $O(10^5$--$10^6)$ samples for reliable generalization. Physics constraints reduce effective hypothesis space complexity through:

\begin{enumerate}
    \item \textbf{\textit{Positivity:}} Concentration outputs $c_i \geq 0$ via softplus activation eliminates half the output space.
    \item \textbf{\textit{Family-specific structure:}} Hierarchical decoder heads enforce sensor physics per family.
    \item \textbf{\textit{Environmental conditioning:}} Cross-sensitivity compensation via learned temperature-humidity-pressure embeddings couples related channels.
    \item \textbf{\textit{Temporal smoothness:}} Total variation regularization enforces physically plausible signal dynamics.
\end{enumerate}

This reduces effective dimensionality from $d$ to $d_{\text{eff}} \ll d$, enabling sample complexity proportional to $d_{\text{eff}}$ rather than $d$.

\subsection{Architecture Variants}
Both PC$^2$DAE variants share the same structural design but differ in channel dimensions:

PC$^2$DAE-Lean (21k parameters): Encoder channels $(20, 28, 20)$, decoder channels $(28, 20)$, environment embedding dimension 12, positivity $\beta=5.0$, dropout 0.1. Optimized for deployment on edge devices (NVIDIA Jetson, Raspberry Pi with accelerator) within UAV SWaP constraints.

PC$^2$DAE-Wide (204k parameters): Encoder channels $(64, 96, 64)$, decoder channels $(96, 64)$, environment embedding dimension 16, positivity $\beta=3.0$, dropout 0.15. Provides maximum representational capacity for offline processing or ground station deployment.

As demonstrated in Section~V, PC$^2$DAE-Lean outperforms the wider variant (67.3\% vs 61.7\% smoothness), suggesting that the stronger inductive bias imposed by reduced capacity, combined with physics constraints, prevents overfitting to noise patterns in the limited training data.

\subsection{Architecture Overview}
\begin{figure*}[!t]
\centering
\includegraphics[width=0.9\textwidth]{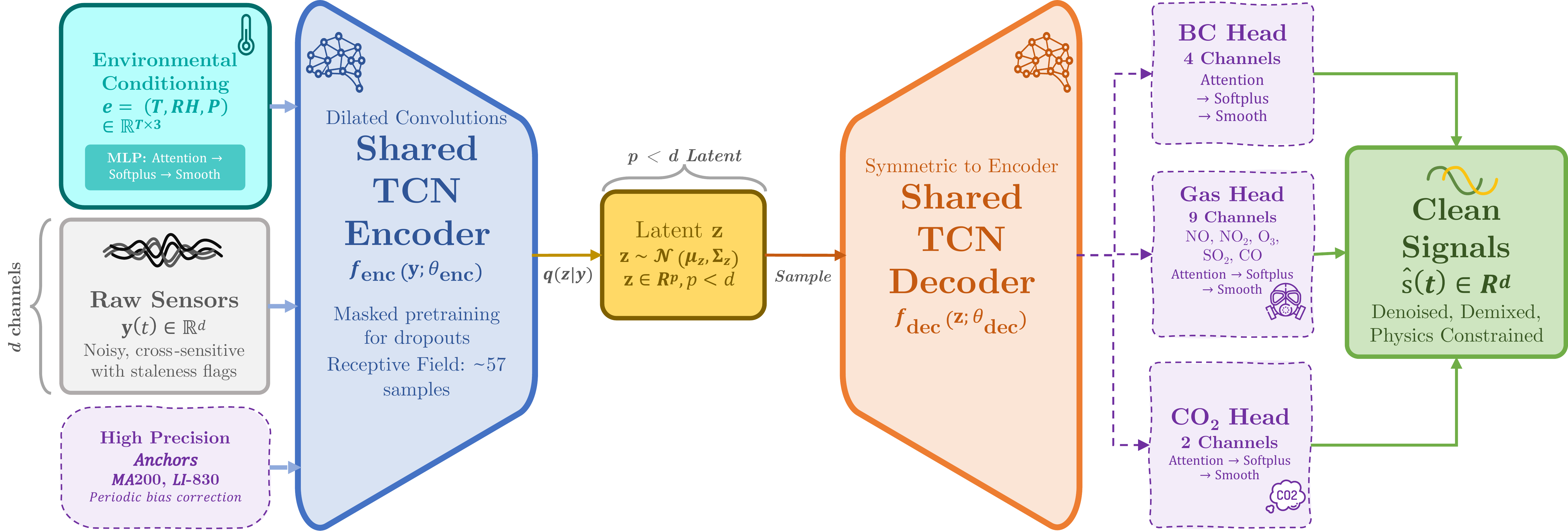}
\caption{PC$^2$DAE architecture for physics-constrained sensor denoising. The shared TCN encoder comprises three dilated 1D convolutional blocks with exponentially increasing dilation factors $(1,2,4)$ yielding $\sim$57-sample receptive field matched to sensor response dynamics. The symmetric decoder feeds into family-specific physics-constrained heads (BC: 4 channels, Gas: 9 channels, CO$_2$: 2 channels), each enhanced with channel attention and learnable temporal smoothing modules. Environmental conditioning ($T$, RH, $P$) modulates decoder outputs to compensate for temperature-dependent drift and humidity cross-sensitivity. The architecture currently executes offboard but is designed for onboard deployment on UAV-class edge hardware.}
\label{fig:pcd2ae}
\end{figure*}

PC$^2$DAE comprises three components: (1) shared TCN encoder, (2) environmental conditioning module, and (3) family-specific physics-constrained decoder heads.
Figure~\ref{fig:pcd2ae} illustrates the complete pipeline.

\subsubsection{Shared Temporal Convolutional Network (TCN) Encoder}
The encoder $f_{\text{enc}}: \mathbb{R}^{T \times C} \to \mathbb{R}^{T \times H}$ maps input with $T$ time steps and $C$ channels to a latent representation with $H$ hidden dimensions, employing dilated temporal convolutions:
\begin{align}
h_1 &= \text{TCNBlock}(x; H_1, k{=}5, d{=}1), \\
h_2 &= \text{TCNBlock}(h_1; H_2, k{=}5, d{=}2), \\
z &= \text{TCNBlock}(h_2; H_3, k{=}5, d{=}4),
\end{align}
where $x$ is the input, $h_1, h_2$ are intermediate hidden states, $z$ is the latent representation, $(H_1, H_2, H_3)$ are $(20, 28, 20)$ for Lean or $(64, 96, 64)$ for Wide, and each TCNBlock is parameterized by output channels, kernel size $k$, and dilation factor $d$. Each TCNBlock consists of dilated one-dimensional convolution (Conv1D), GroupNorm, Exponential Linear Unit (ELU) activation, dropout, and residual connection. The dilation sequence $(1, 2, 4)$ with kernel size $k=5$ yields receptive field $\approx 57$ samples, matching sensor response dynamics ($t_{90} \approx 25$--$80$s at 1\,Hz).

\subsubsection{Environmental Conditioning}
Environmental variables $e = (T_{\text{env}}, H_{\text{env}}, P) \in \mathbb{R}^{T \times 3}$, representing temperature, relative humidity, and pressure respectively, are encoded via a Multi-Layer Perceptron (MLP):
\begin{equation}
e_{\text{embed}} = \text{MLP}(e; 3 \to D_e \to D_e)
\end{equation}
where $D_e = 12$ for Lean or $D_e = 16$ for Wide. This embedding modulates decoder outputs through additive conditioning, compensating for temperature-dependent sensor drift and humidity cross-sensitivity per Eq.~\eqref{eq:ec_comp}.

\subsubsection{Family-Specific Decoder Heads}
Rather than a monolithic decoder, we employ separate physics-constrained heads per sensor family. Each head applies: channel attention $\to$ environmental conditioning $\to$ linear projection $\to$ physics activation $\to$ learnable smoothing.

BC Head (4 channels): Channel attention learns inter-channel relationships (UV/IR correlation), followed by linear projection, positivity via $\text{softplus}(\cdot, \beta)$ where $\beta$ controls the sharpness of the constraint (lower $\beta$ yields softer enforcement), and learnable temporal smoothing with kernel size $k=5$.

Gas Head (9 channels): Channel attention for cross-sensitivity compensation across NO, NO$_2$, O$_3$, SO$_2$, CO. Positivity constraint with adaptive smoothing ($k=5$).

CO$_2$ Head (2 channels): Channel attention, positivity, and learnable smoothing ($k=5$).

\subsubsection{Learnable Temporal Smoothing}
A key innovation is the learnable temporal smoothing module applied after physics constraints:
\begin{equation}
\hat{y}_{\text{smooth}} = \alpha \cdot (K * \hat{y}) + (1-\alpha) \cdot \hat{y}
\end{equation}
where $\hat{y}$ is the physics-constrained output, $K \in \mathbb{R}^{C \times 1 \times k}$ is a per-channel learned smoothing kernel of size $k$ with $\text{softmax}$ normalization (ensuring non-negative weights summing to 1), $*$ denotes convolution, and $\alpha = \sigma(\alpha_0) \in [0,1]$ is a learned blending parameter with $\sigma(\cdot)$ being the sigmoid function and $\alpha_0$ a learnable scalar. This allows the model to learn optimal smoothing strength per channel while preserving physics constraints. Smoothing non-negative values with non-negative weights preserves non-negativity.

\begin{table}[!t]
\centering
\caption{PC$^2$DAE Architecture Specifications (Lean / Wide)}
\label{tab:architecture}
\begin{tabular}{lcccc}
\toprule
Component & Lean & Wide & Kernel & Dilation \\
\midrule
\multicolumn{5}{l}{\textit{Shared Encoder}} \\
TCNBlock 1 & $C \to 20$ & $C \to 64$ & 5 & 1 \\
TCNBlock 2 & $20 \to 28$ & $64 \to 96$ & 5 & 2 \\
TCNBlock 3 & $28 \to 20$ & $96 \to 64$ & 5 & 4 \\
\midrule
\multicolumn{5}{l}{\textit{Shared Decoder}} \\
TCNBlock 4 & $20 \to 28$ & $64 \to 96$ & 5 & 4 \\
TCNBlock 5 & $28 \to 20$ & $96 \to 64$ & 5 & 2 \\
TCNBlock 6 & $20 \to 20$ & $64 \to 64$ & 5 & 1 \\
\midrule
\multicolumn{5}{l}{\textit{Physics Heads}} \\
BC Head & \multicolumn{2}{c}{$\to 4$ channels} & 5 & -- \\
Gas Head & \multicolumn{2}{c}{$\to 9$ channels} & 5 & -- \\
CO$_2$ Head & \multicolumn{2}{c}{$\to 2$ channels} & 5 & -- \\
Env Encoder & $3 \to 12$ & $3 \to 16$ & -- & -- \\
\midrule
Total Params & 21k & 204k & -- & -- \\
\bottomrule
\end{tabular}
\end{table}

\subsection{Physics-Constrained Loss Function}
Training minimizes a multi-objective loss combining reconstruction fidelity with physics regularization:
\begin{equation}
\mathcal{L} = \mathcal{L}_{\text{recon}} + \sum_{g \in \mathcal{G}} \mathcal{L}_{\text{physics}}^{(g)}
\end{equation}
where $\mathcal{G} = \{\text{BC}, \text{Gas}, \text{CO}_2\}$ indexes sensor families.

Reconstruction Loss: Mean Absolute Error (MAE) per family:
\begin{equation}
\mathcal{L}_{\text{recon}} = \sum_{g} \frac{1}{|C_g|} \|\hat{y}^{(g)} - y^{(g)}\|_1
\end{equation}
where $g$ indexes sensor families, $|C_g|$ is the number of channels in family $g$, $\hat{y}^{(g)}$ is the predicted output, $y^{(g)}$ is the target, and $\|\cdot\|_1$ denotes the $L_1$ norm.

Physics Losses: Family-specific constraint penalties:
\begin{align}
\mathcal{L}_{\text{positivity}} &= \lambda_1 \|\text{ReLU}(-\hat{y})\|_1 \\
\mathcal{L}_{\text{smooth}} &= \lambda_2 \sum_t |\hat{y}_{t+1} - \hat{y}_t|
\end{align}
where $\lambda_1, \lambda_2$ are weighting coefficients, $\text{ReLU}(\cdot) = \max(0, \cdot)$ is the Rectified Linear Unit, and $\|\cdot\|_1$ denotes the $L_1$ norm. The subscript $t$ indexes time steps.

For PC$^2$DAE-Lean, we use standard physics weights ($\lambda_{\text{positivity}}=0.1$, $\lambda_{\text{smooth}}=0.01$). For PC$^2$DAE-Wide, reduced weights ($\lambda_{\text{positivity}}=0.01$, $\lambda_{\text{smooth}}=0.005$) allow reconstruction loss to dominate while architectural constraints maintain physics compliance.

\section{Experiments and Results}

\subsection{Baselines}
We compare both PC$^2$DAE variants against five deep learning architectures trained on identical data with comparable training protocols but without physics constraints. These include LSTM-AE (63k params), a bidirectional LSTM encoder-decoder autoencoder with attention; U-Net 1D (820k params), a fully convolutional encoder-decoder with skip connections; Transformer (19k params), employing multi-head self-attention encoder-decoder; CBDAE (49k params), a Contrastive Blind Denoising AE with GRU encoder and NCE loss for noise-invariant representations; and DeSpaWN (1k params), a Deep Spatially-adaptive Wavelet Network with learnable lifting-based wavelet transforms. All baselines use MAE reconstruction loss (CBDAE additionally incorporates NCE loss) without positivity constraints, enabling fair comparison of physics-constrained vs.\ unconstrained learning.

\subsection{Evaluation Metrics}
Smoothness Improvement (\%): Reduction in total variation $\sum_t |y_{t+1} - y_t|$ relative to input.

High-Frequency Noise Reduction (\%): Decrease in spectral power above Nyquist/4.

Physics Violations (\%): Percentage of outputs violating positivity constraint ($\hat{y} < 0$). Concentrations are physically non-negative; violations indicate unphysical predictions.

\subsection{Main Results}
Table~\ref{tab:main_results} presents overall performance across all sensor families. The reported metrics are averaged across BC, Gas, and CO$_2$ sensor families.

\begin{table}[!t]
\centering
\caption{Comparison of PC$^2$DAE Variants with Deep Learning Baselines}
\label{tab:main_results}
\begin{tabular}{lcccc}
\toprule
Model & Params & Smooth.$\uparrow$ & HF Red.$\uparrow$ & Neg.$\downarrow$ \\
\midrule
\textbf{PC$^2$DAE-Lean} & 21k & \textbf{67.3\%} & \textbf{90.7\%} & \textbf{0.0\%} \\
\textbf{PC$^2$DAE-Wide} & 204k & 61.7\% & 82.2\% & \textbf{0.0\%} \\
DeSpaWN~\cite{doi:10.1073/pnas.2106598119} & \textbf{1k} & 47.7\% & 84.1\% & 15.3\% \\
LSTM-AE & 63k & 11.9\% & 59.6\% & 22.2\% \\
CBDAE~\cite{langaricaContrastiveBlindDenoising2023} & 49k & 10.9\% & 60.5\% & 22.4\% \\
U-Net 1D & 820k & $-$0.7\% & 34.0\% & 20.8\% \\
Transformer & 19k & $-$28.9\% & $-$35.7\% & 23.1\% \\
\bottomrule
\end{tabular}
\end{table}

PC$^2$DAE-Lean achieves the best overall performance with 67.3\% smoothness improvement and 90.7\% high-frequency noise reduction, while maintaining zero physics violations. Notably, the lean variant outperforms the wider PC$^2$DAE-Wide by +5.6\% smoothness and +8.5\% HF reduction despite having 10$\times$ fewer parameters. This counterintuitive result suggests that stronger inductive bias from reduced capacity, combined with physics constraints, prevents overfitting to noise patterns.

Both PC$^2$DAE variants achieve zero physics violations, the only models to do so. The negative smoothness/HF metrics for Transformer ($-$28.9\%, $-$35.7\%) indicate this model amplifies noise rather than suppresses it. With only $\sim$8k samples, the attention-based architecture fails to learn meaningful denoising patterns. Similarly, U-Net (820k params) and CBDAE (49k params) achieve only marginal or negative smoothness, demonstrating that high capacity without physics constraints leads to overfitting.

All unconstrained baselines produce 15--23\% physics violations (negative concentration outputs), a fundamental failure mode for sensor data that both PC$^2$DAE variants eliminate through architectural constraints.

\subsection{Per-Family Analysis}

Tables~\ref{tab:per_family_smooth}--\ref{tab:per_family_neg} disaggregate performance by sensor family, revealing family-specific challenges.

\begin{table}[!t]
\centering
\caption{Per-Family Smoothness Improvement (\%)}
\label{tab:per_family_smooth}
\begin{tabular}{lccc}
\toprule
Model & BC & Gas & CO$_2$ \\
\midrule
\textbf{PC$^2$DAE-Lean} & \textbf{65.1} & \textbf{74.7} & 62.2 \\
\textbf{PC$^2$DAE-Wide} & 61.0 & 61.2 & \textbf{62.9} \\
DeSpaWN & 42.6 & 69.0 & 31.5 \\
LSTM-AE & 19.7 & 23.2 & $-$7.3 \\
CBDAE & 16.9 & 19.5 & $-$3.6 \\
U-Net 1D & 2.6 & 4.2 & $-$9.0 \\
Transformer & $-$53.9 & $-$27.4 & $-$5.3 \\
\bottomrule
\end{tabular}
\end{table}

\begin{table}[!t]
\centering
\caption{Per-Family High-Frequency Noise Reduction (\%)}
\label{tab:per_family_hf}
\begin{tabular}{lccc}
\toprule
Model & BC & Gas & CO$_2$ \\
\midrule
\textbf{PC$^2$DAE-Lean} & 95.3 & 93.0 & \textbf{83.9} \\
\textbf{PC$^2$DAE-Wide} & \textbf{95.5} & 71.7 & 79.5 \\
DeSpaWN & 78.0 & \textbf{95.9 }& 78.5 \\
LSTM-AE & 73.8 & 69.3 & 35.6 \\
CBDAE & 71.0 & 71.4 & 39.2 \\
U-Net 1D & 39.0 & 25.0 & 38.1 \\
Transformer & $-$45.7 & $-$63.8 & 2.5 \\
\bottomrule
\end{tabular}
\end{table}

\begin{table}[!t]
\centering
\caption{Per-Family Physics Violations (\% Negative Outputs)}
\label{tab:per_family_neg}
\begin{tabular}{lccc}
\toprule
Model & BC & Gas & CO$_2$ \\
\midrule
\textbf{PC$^2$DAE-Lean} & \textbf{0.0} & \textbf{0.0} & \textbf{0.0} \\
\textbf{PC$^2$DAE-Wide} & \textbf{0.0} & \textbf{0.0} & \textbf{0.0} \\
DeSpaWN & 25.7 & 20.1 & \textbf{0.0} \\
LSTM-AE & 41.3 & 25.3 & \textbf{0.0} \\
CBDAE & 40.5 & 26.8 & \textbf{0.0} \\
U-Net 1D & 37.1 & 25.4 & \textbf{0.0} \\
Transformer & 40.0 & 29.3 & \textbf{0.0} \\
\bottomrule
\end{tabular}
\end{table}

The BC family shows the largest violation rates: All baselines produce 26--41\% negative outputs on BC channels, which frequently approach zero during background periods. Both PC$^2$DAE variants' softplus constraint guarantees valid outputs regardless of concentration level.

Gas family highlights cross-sensitivity: The 9-channel Gas family exhibits complex cross-sensitivity patterns (e.g., O$_3$ affecting NO$_2$ response). PC$^2$DAE-Lean achieves 74.7\% smoothness with zero violations, substantially outperforming PC$^2$DAE-Wide (61.2\%) on this challenging family. Baselines produce 20--29\% violations due to cross-sensitivity artifacts that drive apparent concentrations negative.

CO$_2$ presents different challenges: All models achieve 0\% violations on CO$_2$ because ambient levels ($\sim$420ppm) ensure positive values. However, baselines show negative smoothness improvement ($-$3.6\% to $-$9.0\%), indicating they introduce additional noise into this relatively stable signal. Both PC$^2$DAE variants achieve $\sim$62\% smoothness improvement and $\sim$80\% HF reduction.

PC$^2$DAE-Lean excels on Gas channels: The lean variant achieves 93.0\% HF reduction on Gas (vs.\ Wide's 71.7\%), demonstrating that reduced capacity with physics constraints can outperform larger unconstrained models on complex multi-channel data. Table~\ref{tab:traditional} compares PC$^2$DAE-Lean against classical signal processing methods under synthetic noise injection (medium level: noise standard deviation $\sigma=0.05$), enabling direct measurement of denoising capability.

\begin{table}[!t]
\centering
\caption{Comparison with Traditional Signal Processing Methods}
\label{tab:traditional}
\begin{tabular}{lccc}
\toprule
Method & MAE Improv.$\uparrow$ & SNR Improv.$\uparrow$ & Neg.$\downarrow$ \\
\midrule
\textbf{PC$^2$DAE Lean (Ours)} & \textbf{+46.4\%} & \textbf{+9.2 dB} & \textbf{0.0\%} \\
Raw (no filter) & 0.0\% & 0.0 dB & 38.5\% \\
Moving Avg (w=5) & $-$7.9\% & +4.6 dB & 38.2\% \\
Moving Avg (w=11) & $-$38.5\% & +2.3 dB & 37.4\% \\
Wavelet Denoising & $-$10.7\% & +1.1 dB & 37.8\% \\
Savitzky-Golay & $-$14.4\% & +4.3 dB & 38.5\% \\
Kalman Filter & $-$61.4\% & $-$0.1 dB & 37.6\% \\
\bottomrule
\end{tabular}
\end{table}

Traditional filters achieve negative MAE improvement because they distort signals more than they denoise. Fixed-bandwidth filters blur sharp but genuine concentration gradients (plume edges), while PC$^2$DAE learns to distinguish real transients from noise. Under high noise ($\sigma=0.10$), PC$^2$DAE's advantage increases to +75.7\% MAE improvement.

\subsection{Qualitative Signal Analysis}

\begin{figure*}[!t]
\centering
\subfloat[BC signal reconstruction]{\includegraphics[width=\columnwidth]{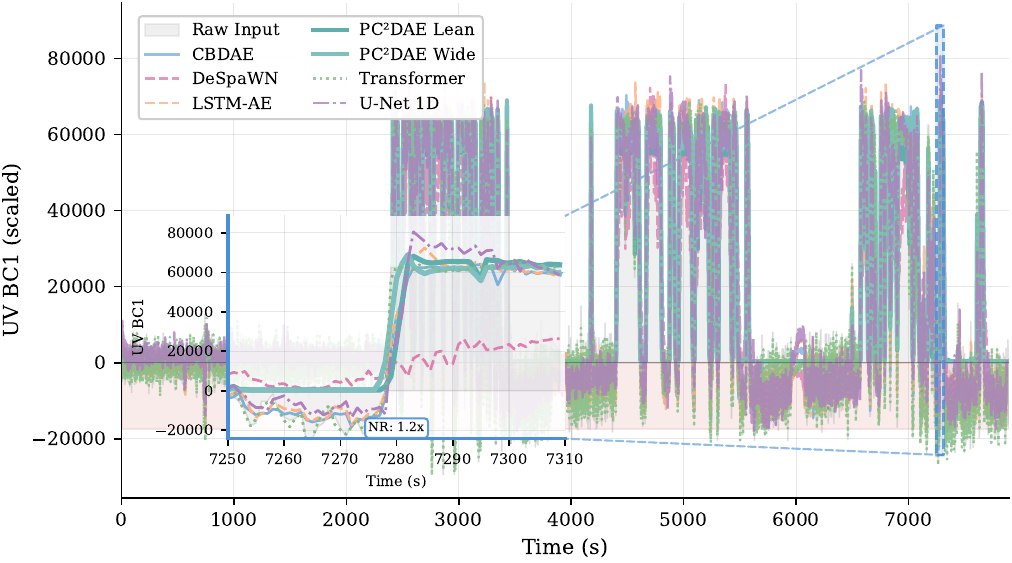}\label{fig:signal_bc}}
\hfil
\subfloat[Gas sensor signal reconstruction]{\includegraphics[width=\columnwidth]{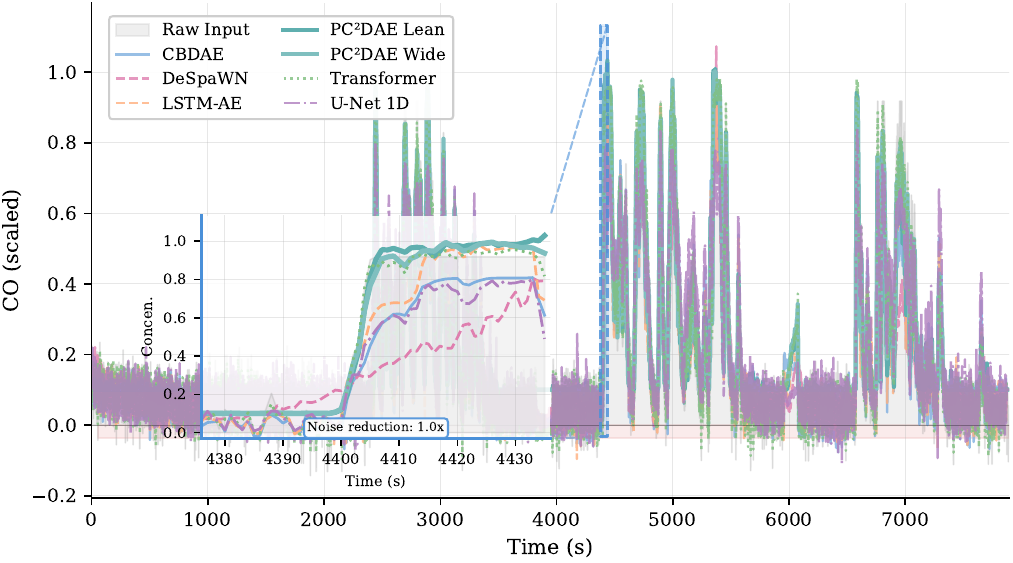}\label{fig:signal_gas}}
\caption{Signal reconstruction comparison for Black Carbon and Gas sensor families. (a) Black Carbon UV channel: Raw input (gray shaded) exhibits characteristic high-frequency noise. PC$^2$DAE-Lean (teal) achieves 65.1\% smoothness improvement and 95.3\% HF noise reduction while maintaining strict positivity. LSTM-AE (orange dashed) achieves lower smoothness (19.7\%) and produces 41.3\% physics violations (negative outputs). Inset shows detail of noise suppression. (b) Gas sensor CO channel: PC$^2$DAE-Lean achieves 74.7\% smoothness and 93.0\% HF reduction with zero violations, substantially outperforming PC$^2$DAE-Wide (61.2\%, 71.7\%). Unconstrained baselines produce 20--29\% negative outputs due to cross-sensitivity artifacts.}
\label{fig:signal_comparison}
\end{figure*}

Figure~\ref{fig:signal_comparison} illustrates denoised outputs on representative test segments for the BC and Gas sensor families, the most critical measurements for wildfire monitoring.

\begin{figure}[!t]
\centering
\includegraphics[width=\columnwidth]{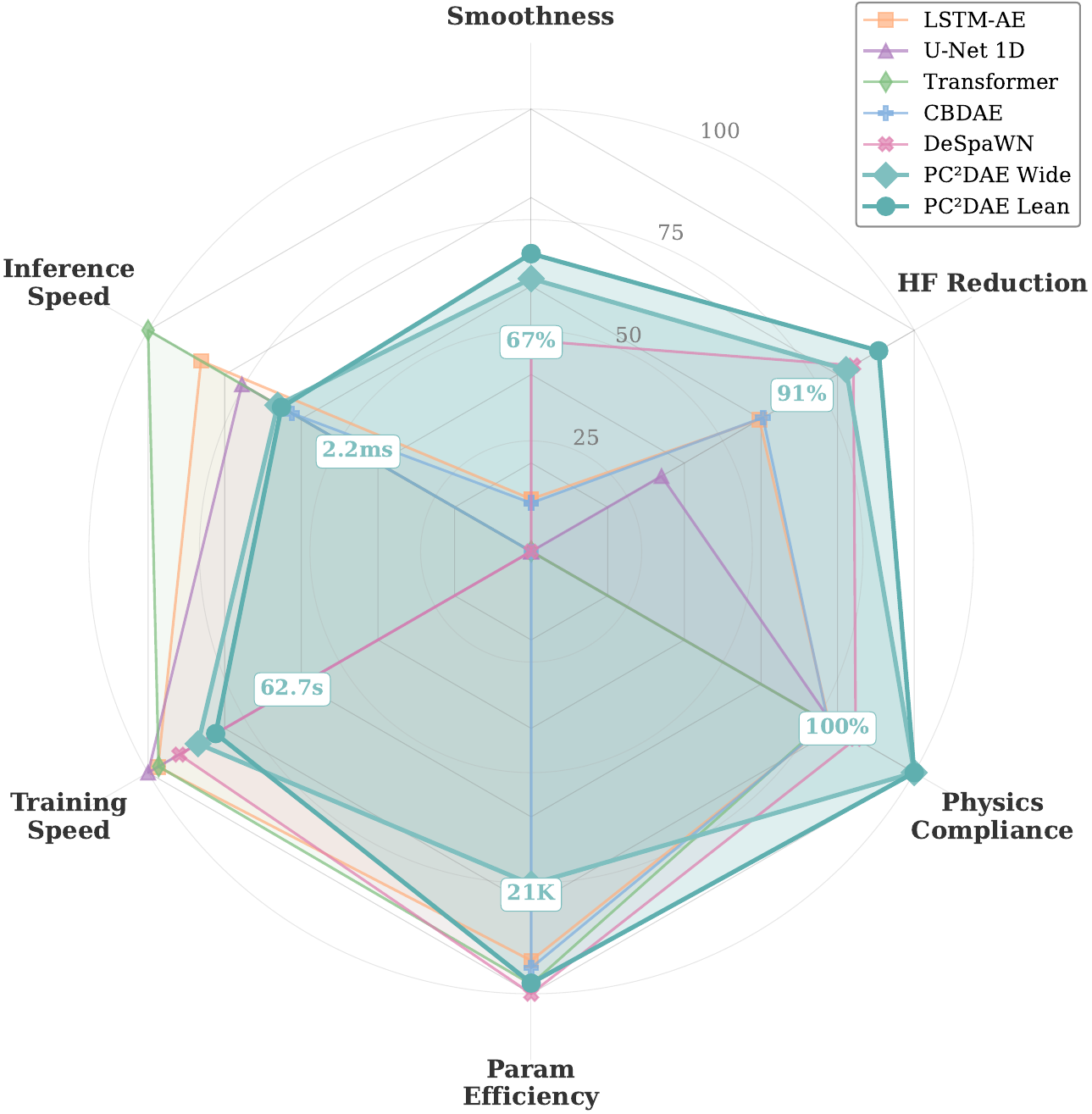}
\caption{Multi-dimensional comparison across six evaluation axes: Smoothness, HF Reduction, Physics compliance (100 $-$ \% violations), Parameter Efficiency (inverse of parameter count), Training Speed, and Inference Speed. PC$^2$DAE-Lean (dark teal) achieves 100\% physics compliance while maximizing denoising performance with 21k parameters, 62.7s training time, and 2.2ms inference per window. PC$^2$DAE-Wide (light teal) also achieves perfect physics compliance. All unconstrained baselines fail on physics compliance (77--85\%), producing physically impossible negative concentrations.}
\label{fig:radar_comparison}
\end{figure}

Physics violations under low concentrations: During calibration periods and background measurements, baselines drift into negative concentration values. On BC, all baselines produce 26--41\% violations. Both PC$^2$DAE variants' softplus positivity constraint guarantees valid outputs regardless of concentration level. Cross-sensitivity artifacts: Gas sensors exhibit inter-channel interference. While LSTM-AE learns channel correlations, it cannot enforce positivity when cross-sensitivity drives apparent concentrations below zero. PC$^2$DAE's channel attention mechanism learns these correlations while the physics head maintains validity. Figure~\ref{fig:radar_comparison} provides a holistic view across five evaluation dimensions. Both PC$^2$DAE variants achieve perfect physics compliance (100\%), distinguishing them from all baselines. PC$^2$DAE-Lean additionally maximizes parameter efficiency while achieving the highest smoothness and HF reduction. The radar plot highlights that while DeSpaWN achieves reasonable smoothness (47.7\%) with minimal parameters (1k), it fundamentally fails on physics compliance (85\% vs 100\%), a binary requirement for valid sensor data.

\subsection{Why Lean Outperforms Wide}

The counterintuitive finding that PC$^2$DAE-Lean (21k params) outperforms PC$^2$DAE-Wide (204k params) merits discussion. We hypothesize several contributing factors:

Capacity-constraint synergy: With only $\sim$8k training samples, a 204k parameter model has sufficient capacity to memorize noise patterns. The lean variant's reduced capacity forces generalization, while physics constraints ensure the learned mapping remains physically valid.

Regularization through architecture: The lean architecture implicitly regularizes by limiting representational capacity. This architectural regularization complements the explicit physics constraints, whereas the wide model requires careful loss weighting (10$\times$ reduced physics weights) to balance reconstruction and physics objectives.

Gradient dynamics: Smaller models often exhibit more stable gradient flow during training. The lean variant converged with standard physics loss weights, while the wide variant required hyperparameter tuning to avoid mode collapse.

\section{Conclusion}

We presented PC$^2$DAE, a hierarchical physics-constrained denoising autoencoder in two variants: PC$^2$DAE-Lean (21k parameters) and PC$^2$DAE-Wide (204k parameters). Evaluation on 7,894 training samples across three sensor families (BC, Gas, CO$_2$) demonstrates that PC$^2$DAE-Lean suggesting that stronger inductive bias from reduced capacity, combined with physics constraints, prevents overfitting while maintaining zero physics violations. Notably, the lean variant outperforms the wider model (+5.6\% smoothness, +8.5\% HF reduction), suggesting that a stronger inductive bias from reduced capacity, combined with physics constraints, prevents overfitting in data-scarce regimes. The hierarchical architecture with family-specific decoder heads enables principled sensor selection based on data quality, excluding sensor families with insufficient signal-to-noise ratios to prevent overfitting to noise. Key innovations include learnable temporal smoothing modules and channel attention mechanisms enabling adaptive, per-channel denoising, while architectural physics constraints via softplus activation guarantee physically valid outputs by construction. Compared to five unconstrained deep learning baselines, both PC$^2$DAE variants are the only models achieving zero physics violations, while all baselines produce 15-23\% negative outputs. The design principle---that when data is scarce, one should embed physics in architecture, not just in loss functions, and prefer smaller models with stronger inductive bias, offers a template for physics-informed machine learning in data-limited sensing applications.


\bibliography{cleaned}
\bibliographystyle{ieeetr}

\end{document}